%% file: PaperForReview.tex
\documentclass[10pt,twocolumn,letterpaper]{article}

\usepackage[pagenumbers]{cvpr} 

\usepackage{graphicx}
\usepackage{enumitem} 
\usepackage{amsmath}
\usepackage{amssymb}
\usepackage{booktabs}

\usepackage[pagebackref,breaklinks,colorlinks]{hyperref}

\usepackage[capitalize]{cleveref}
\crefname{section}{Sec.}{Secs.}
\Crefname{section}{Section}{Sections}
\Crefname{table}{Table}{Tables}
\crefname{table}{Tab.}{Tabs.}

\usepackage[accsupp]{axessibility}  
\usepackage{multirow}
\usepackage{rotating}
\usepackage{adjustbox}
\def\eg{\emph{e.g}\onedot} 
\def\ie{\emph{i.e}\onedot} 
\usepackage{mathtools}
\usepackage{textcomp}
\usepackage{gensymb}
\usepackage{enumitem}

\usepackage[dvipsnames]{xcolor}
\newcommand*{\red}{\textcolor{red}}
\newcommand*{\green}{\textcolor{green}}

\newcommand{\myparagraph}[1]{\vspace{4pt}\noindent\textbf{#1}}

\def\cvprPaperID{15} 
\def\confName{CVPR}
\def\confYear{2023}

\usepackage{amssymb}
\usepackage{pifont}
\newcommand{\cmark}{\ding{51}}
\newcommand{\xmark}{\ding{55}}
\newcommand\RX{\text{\red{\xmark}}}
\newcommand\GT{\text{\green{\cmark}}}

\usepackage{float}

\makeatletter
\@namedef{ver@everyshi.sty}{}
\makeatother
\usepackage{tikz}

\begin{document}

\title{Are Local Features All You Need for Cross-Domain Visual Place Recognition?}

\author{Giovanni Barbarani, Mohamad Mostafa, Hajali Bayramov,\\Gabriele Trivigno, Gabriele Berton, Carlo Masone and Barbara Caputo\\
Politecnico di Torino
\\
{\tt\small [giovanni.barbarani, mohamad.mostafa, hajali.bayramov]@studenti.polito.it} \\{\tt\small[gabriele.trivigno, gabriele.berton, carlo.masone, barbara.caputo]@polito.it}
}
\maketitle

\begin{abstract}
Visual Place Recognition is a task that aims to predict the coordinates of an image (called query) based solely on visual clues. Most commonly, a retrieval approach is adopted, where the query is matched to the most similar images from a large database of geotagged photos, using learned global descriptors.
Despite recent advances, recognizing the same place when the query comes from a significantly different distribution is still a major hurdle for state of the art retrieval methods. Examples are heavy illumination changes (e.g. night-time images) or substantial occlusions (e.g. transient objects).  
In this work we explore whether re-ranking methods based on spatial verification can tackle these challenges, following the intuition that local descriptors are inherently more robust than global features to domain shifts. To this end, we provide a new, comprehensive benchmark on current state of the art models.
We also introduce two new demanding datasets with night and occluded queries, to be matched against a city-wide database.
Code and datasets are available at 
{\small{\url{https://github.com/gbarbarani/re-ranking-for-VPR}}}.
\end{abstract}


\input{chapters/ch1_introduction}

\input{chapters/ch2_related_work}

\input{chapters/ch3_dataset}

\input{chapters/ch4_experiments}

\input{chapters/ch5_conclusion}

{\small
\bibliographystyle{ieee_fullname}
\bibliography{egbib}
}

\end{document}

%% file: chapters/ch1_introduction.tex
\section{Introduction}
\label{sec:introduction}

The task of Visual Place Recognition (VPR) aims to answer the question ``\textit{Where was this picture taken?}''. 
In the literature the most popular approach is to cast the task as an image retrieval problem, where a given query is localized via comparison to a previously collected database of geotagged images \cite{Zaffar_2021_vprbench, Berton_2022_benchmark, Arandjelovic_2018_netvlad, Kim_2017_crn, Liu_2019_sare, Ge_2020_sfrs, Berton_2022_cosPlace, Alibey_2023_mixvpr, Alibey_2022_gsvcities, Paolicelli_2022_SegVPR, Zhu_2023_r2former, Leyvavallina_2021_gcl}, and the query is considered correctly localized of its ground truth position is less then 25 meters away form the prediction.
VPR can be used as a first step before more precise visual localization, and can find multiple applications in fields like autonomous driving, SLAM and augmented reality.
Given these applications, the task is usually performed in large-scale outdoor scenarios, for which the database is collected in an automated fashion, typically via Street View data \cite{Torii_2015_pitts250k, Torii_2018_tokyo247, Berton_2021_svox, Berton_2022_cosPlace}, which ends up being made up of mostly day-time images.
On the other hand, the queries that a real-world VPR system receives once deployed may be subject to high appearances changes, due to night-time images, occlusions, critical meteorological conditions. This domain shift between queries and database still represents a major challenge in the literature \cite{Torii_2018_tokyo247, Asha_2019_todaygan, Wang_2019_DA_for_VPR, Berton_2021_svox, Warburg_2020_msls, Ibrahimi_2021_insideout_vpr, Yildiz_2022_AmsterTime, Piasco_2018_survey, Zhang_2021_survey, Masone_2021_survey}.

\input{figures/plot_r1_time}

To improve results and address these issues, a number of previous works noted that local features \cite{Noh_2017_delf, Cao_2020_delg, Lee_2022_cvnet, Wang_2022_TransVPR} and image matching \cite{Revaud_2019_r2d2, Dusmanu_2019_D2Net, Sun_2021_loftr, Sarlin_2020_superglue} methods are inherently more robust to domain shifts , and that these can be used to re-rank a set of candidates (usually through spatial verification) provided through image retrieval methods, leading to large improvements in results \cite{Hausler_2021_patch_netvlad, Wang_2022_TransVPR}.



Our work aims to quantify the effectiveness of these methods that provide a matching score between two images, when applied to re-rank the top-N candidates of a retrieval module for VPR.
Despite the recent interest in this category of methods, they are yet to be studied in the VPR setting, and previous work only compare a small number of these methods \cite{Hausler_2021_patch_netvlad, Wang_2022_TransVPR}.
Furthermore, previous comparisons of such methods in a VPR setting are hindered by the use of different underlying retrieval methods: for example, in \cite{Wang_2022_TransVPR} the authors use DELG, Patch-NetVLAD, TransVPR and SuperGlue to re-rank a shortlist of candidates provided by different retrieval methods, making it difficult to understand whether better results are granted by re-ranking or the retrieval module.

To establish which methods are most suited for re-ranking in real-world VPR, we perform an extensive benchmarks of existing image matching pipelines with a focus on the domain shift problem.
In particular, we focus on creating fair benchmarking conditions, by providing all the re-ranking methods with the same pool of candidates to score.
Where possible, we use the same backbone for local feature extraction, and use the same hardware to carry out extensive efficiency evaluation.
Our benchmark reveals that even highly challenging datasets can be nearly solved by combining SOTA retrieval and re-ranking methods (\eg on Tokyo night, which uses only the night queries of Tokyo 24/7, we achieved a Recall@1 $> 95\%$).

We therefore propose two new challenging datasets, in order to provide a stimulating and challenging benchmark to foster future research.
In the creation of these two datasets we focused on the two most challenging domains for VPR: the first has night-time queries, whereas the second has queries with heavy occlusions due to dynamic objects (\eg vehicles and pedestrians).
For both datasets, we collected (and manually verified) queries from Flickr, whereas as a database we use the San Francisco eXtra Large (SF-XL) dataset.

\input{figures/reranking_pipeline}

\textbf{Our contributions} can be summarized in the following points:
\begin{itemize}[noitemsep,topsep=1pt]

    \item We propose two new query sets to allow to evaluate the performance on night-time and occluded images against a city-wide database. 
    Both query sets have been collected from Flickr and manually curated.
    
    \item We construct a benchmark to explore the applicability of spatial verification techniques for re-ranking in VPR. We create comparable setups to isolate the performances of the tested methods, quantify their gains with respect to the state of the art in VPR.
    
    \item We find that re-ranking methods are able to greatly improve the results over commonly used retrieval methods, and we observe that there is no clear winning solution, as different scenarios require different methods.

\end{itemize}

%% file: figures/plot_r1_time.tex
\begin{figure}
    \centering
    \includegraphics[width=\columnwidth]{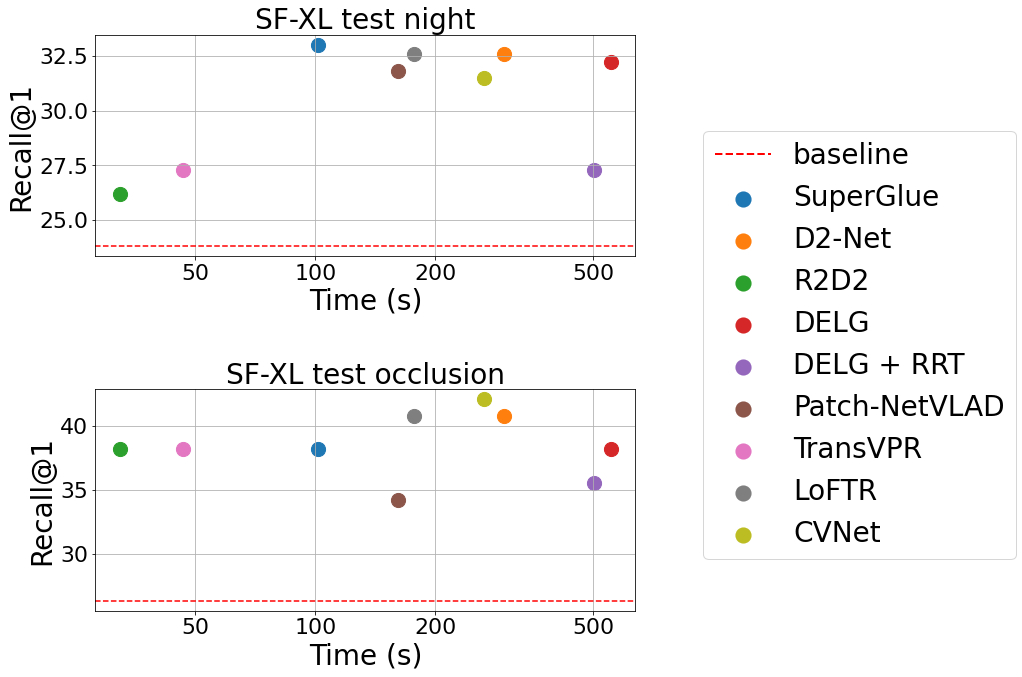}
    \caption{\textbf{Plot showing the Recall@1 and latency for different methods on multiple datasets.} Latency is to re-rank 100 candidates of a single query, considering local features extraction to be performed online.
    We can see that there is no single method that outperforms all others on all scenarios, and the ideal choice of a re-ranking method for a VPR system depends on multiple factors, such as time requirements and expected domain shifts. 
    }
    \label{fig:plot_r1_time}
\end{figure}

%% file: figures/reranking_pipeline.tex
\begin{figure}
    \centering
    \includegraphics[width=\columnwidth]{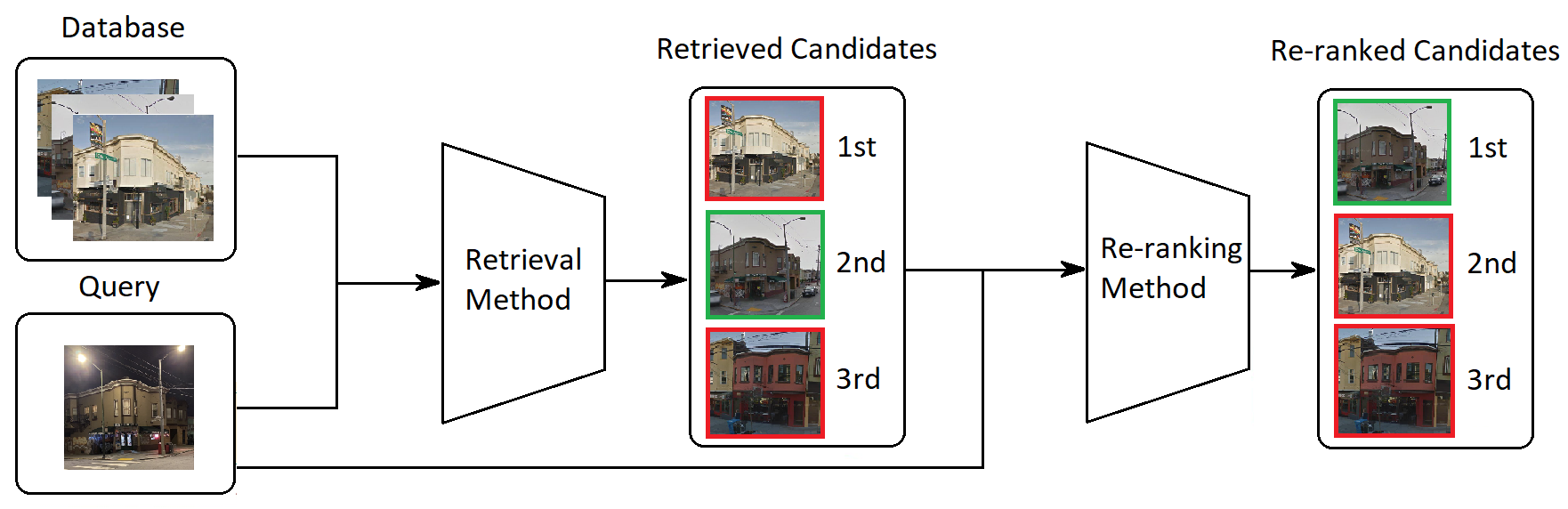}
    \caption{\textbf{Overview of the re-ranking pipeline.} First, a retrieval method performs a similarity search on the global descriptors extracted from query and database to output a set of top-k candidates. Then re-ranking is applied to refine the retrieved candidates.}
    \label{fig:reranking_pipeline}
    \vspace{-3mm}
\end{figure}

%% file: chapters/ch2_related_work.tex
\section{Related Work}
\label{sec:related_work}

\myparagraph{Visual Place Recognition through Image Retrieval}

Image Retrieval is the most common way to approach the task of Visual Place Recognition.
A neural network is used to extract global descriptors from the query and database images, and then a kNN is performed to find the matches to the query.
Among the proposed global extractors, NetVLAD \cite{Arandjelovic_2018_netvlad}, a deep learning successor of VLAD \cite{Jegou_2011_vlad}, established a milestone in the field of VPR.
NetVLAD led to the birth of a large number of work that proposed improvements to it, commonly trained with variants of the weakly supervised triplet loss: among them we note ApaNet \cite{Zhu_2018_apanet}, CRN \cite{Kim_2017_crn}, SARE \cite{Liu_2019_sare}, SFRS \cite{Ge_2020_sfrs}, AppSVR \cite{Peng_2021_appsvr} and SralNet \cite{Peng_2021_sralNet}.
Such methods have been recently outperformed by models that do not rely on NetVLAD, use smaller descriptors, and propose new training techniques to scale to large training sets: the most notable examples of this trend are CosPlace \cite{Berton_2022_cosPlace}, Conv-AP \cite{Alibey_2022_gsvcities} and MixVPR \cite{Alibey_2023_mixvpr}.

A separate line of works \cite{Wang_2019_DA_for_VPR, Asha_2019_todaygan, Berton_2021_svox} propose to explicitly tackle the domain shift issue through domain adaptation, although such methods are by nature focusing on a single domain, lacking generalization capabilities.

While retrieval methods have been covered by a number of benchmarks throughout the years \cite{Sattler_2018_aachen_daynight, Pion_2020_benchmark, Zaffar_2021_vprbench, Berton_2022_benchmark}, no benchmark has focused on the possibilities that re-ranking method offers and the computational trade-offs they entail.

\input{tables/dataset_table}
\myparagraph{Local features for spatial verification}

Local features-based spatial verification represents an established paradigm that has been applied to several computer vision tasks, ranging from structure from motion (SfM) \cite{Li_2018_megadepth, Schnberger_2016_colmap, Liu_2017_sfm}, simultaneous localization and mapping (SLAM) \cite{Sarlin_2020_superglue, Hausler_2021_patch_netvlad, Cummins_2011_fabmap} and visual localization \cite{Li_2018_megadepth, Sattler_2018_aachen_daynight, Sun_2021_loftr, Torii_2018_tokyo247}.
For many years hand-crafted feature extractors have represented a remarkably strong baseline \cite{Lowe_2004_sift, Bay_2008_surf}, whereas pioneering learning-based methods showed large margin for improvements under perspective and lightning changes \cite{Yi_2016_lift, Detone_2017_magic}.
In recent years, we have witnessed a flourishing literature on learnable detectors and descriptors exploiting local features for pose and homography estimation \cite{detone2017toward, DeTone_2018_superpoint, Dusmanu_2019_D2Net, Revaud_2019_r2d2, Sarlin_2020_superglue, Sun_2021_loftr, Jin_2021_matchingbench}. 
Although these methods are not specifically designed for retrieval, they can be naturally used to re-rank retrieval candidates by assigning a higher similarity to pairs of images that share more matches across local features.

However, methods trained for outdoor image matching 
can be less robust to dynamic objects (\eg they may match the cars between two images instead of the buildings), than other methods that were specifically trained for image retrieval and re-ranking \cite{Hausler_2021_patch_netvlad, Cao_2020_delg}.

Generally, local features are represented as pairs of keypoint (\ie the pixel coordinate of the feature) and descriptor (a fixed size vector).
Given a pair of images, the local features are then cross-matched to find pairs of keypoints across the two images, in a procedure known as spatial verification.
While this step is usually performed with heuristics like RANSAC \cite{fischler1981random}, data-driven approaches like SuperGlue \cite{Sarlin_2020_superglue} have been proposed for the task. 

SuperGlue uses graph neural networks to learn data-dependent priors on matches given two sets of keypoints as an input.
This approach has been subsequently generalized by LoFTR \cite{Sun_2021_loftr}, which removes the dependence from an underlying detector exploiting cross-attention transformers for directly selecting keypoints matches among an image pair.

Some of these methods have been evaluated on outdoor datasets that are not commonly used in the VPR literature. Some examples are Oxford5k \cite{Philbin_2007_oxford5k} and Paris6k \cite{Philbin_2008_paris6k}, not suited for VPR since they do not provide a dense database (densely covering a given area), and others, like Aachen \cite{Sattler_2012_aachen, Sattler_2018_aachen_daynight}, cover only a small area of a city, and are mainly used for pose estimation.
The same consideration holds for Madrid Metropolis, Gendarmenmarkt and Tower of London, proposed in \cite{Wilson_2014_1dsfm}.

\myparagraph{Local features for re-ranking}

Local features have been explored also for image retrieval, mainly with purpose of re-ranking the shortlist of top-N candidates proposed by the retrieval module \cite{Hausler_2021_patch_netvlad, Fuwen_2021_reranking_transformers, Wang_2022_TransVPR, Cao_2020_delg}. 
To this end, the matching algorithm needs to be turned in a scoring algorithm, commonly using the number of matches found in an image pair as a proxy of confidence.
In other cases, instead of an explicit re-ranking step, local features are used either to enhance global descriptors or to match directly reference images \cite{Yang_2021_dolg, Noh_2017_delf, Weinz_2022_fire, Tolias_2020_how}.
DELG \cite{Cao_2020_delg} uses a global extractor trained with a large margin cosine loss, coupled with local features refined following unsupervised criteria for discriminativeness and reliability.
Patch-NetVLAD \cite{Hausler_2021_patch_netvlad} obtains a set of dense local features performing the VLAD aggregation on local patches, while TransVPR \cite{Wang_2022_TransVPR} is based on a transformer architecture that selects a subset of patches through its multi-scale attention maps. These last two methods were designed for VPR.
The majority of re-ranking methods exploit RANSAC, using the number of inliers to assign a score to the candidates \cite{Noh_2017_delf, Cao_2020_delg, Hausler_2021_patch_netvlad, Wang_2022_TransVPR}. 

Recently, researchers have been explored end-to-end learnable architectures able to estimate a similarity score between pairs of images, as alternative to RANSAC. In \cite{Fuwen_2021_reranking_transformers}, the authors feed the local and global features from DELG to a transformers architecture and then cast the problem as a binary classification task.
Similarly, CVNet \cite{Lee_2022_cvnet} builds a pyramid of 4D correlation maps from the feature maps of a CNN. The 4D maps are then reduced to a similarity score through 4D convolutions.
Unlike many methods for homography estimation, the algorithms mentioned in this section are trained without patch-level supervision.

%% file: tables/dataset_table.tex
\begin{table*}[t]
\begin{adjustbox}{width=\linewidth}
\centering
\begin{tabular}{lccccc}
\toprule
Database & Database source & Database \# images & Query set & Queries source & \# queries \\
\hline
Tokyo 24/7 \cite{Torii_2018_tokyo247} & Google StreetView & 75k & Tokyo night (night queries from Tokyo 24/7) & Collected with a smartphone by \cite{Torii_2018_tokyo247} & 105 \\
\hline
SVOX \cite{Berton_2021_svox} & Google StreetView &  17k  & SVOX Night & Oxford RobotCar & 823 \\
\hline
\multirow{4}{*}{SF-XL \cite{Berton_2022_cosPlace}} & \multirow{4}{*}{Google StreetView} & \multirow{4}{*}{2.8M} & SF-XL test v1 & Flickr & 1000 \\
& & &  SF-XL test v2 & Collected with a smartphone by \cite{Chen_2011_san_francisco} & 598 \\ 
& & &  SF-XL test night (\textbf{ours}) & Flickr & 466 \\
& & &  SF-XL test occlusion (\textbf{ours}) & Flickr & 76 \\
\bottomrule
\end{tabular}
\end{adjustbox}
\caption{\textbf{Summary of the datasets considered in our experiments.} The table reports the raw data sources, every author has cleaned and processed them in customized way, refer to their papers for the details. In general streetview panoramas have been cropped in patches and turned in multiples suitable references for the databases. Flickr queries have been filtered and manually checked for positive references. Oxford RobotCar \cite{Maddern_2017_robotCar} data have been collected and processed with a modality analogues to streetview panoramas, although Tokyo 24/7 \cite{Torii_2018_tokyo247} queries have been collected with smartphone devices they are scenes compatible with a moving vehicle point of view. While Flickr and San Francisco Landmark \cite{Chen_2011_san_francisco} data contains a broader range of point of views and camera types.}
\label{tab:dataset}
\vspace{-3mm}
\end{table*}

%% file: chapters/ch3_dataset.tex
\section{Dataset}
\label{sec:dataset}

\myparagraph{Previous datasets}


Several datasets with query splits that explicitly aim to measure the domain shift adaptation capability have been proposed in the VPR literature \cite{Torii_2018_tokyo247, Berton_2021_svox, Maddern_2017_robotCar, Berton_2022_cosPlace}.

Among these, Tokyo 24/7 \cite{Torii_2018_tokyo247} presents three queries splits taken from different times of the day, respectively daytime, sunset, night, providing a widely used dataset for cross-domain VPR.
While Tokyo 24/7 provides a well curated dataset and is widely used in literature \cite{Arandjelovic_2018_netvlad, Liu_2019_sare, Ge_2020_sfrs, Liu_2020_densernet}, we argue that it is quite limited in size: there are 105 queries per domain and a database of 75k images, covering just a small part of the city.

In \cite{Berton_2021_svox} presented SVOX, a cross-domain dataset with database from Google StreetView. The queries come from RobotCar \cite{Maddern_2017_robotCar} and belong to a number of domains, namely snow, rain, sun, night and overcast.
SVOX uses a database smaller than Tokyo 24/7's, and it only provides frontal-view images, \ie with the facing straight along the road.

Another cross-domain dataset is San Francisco eXtra Large (SF-XL) \cite{Berton_2022_cosPlace}, with a large-scale database that covers the whole city of San Francisco with 2.8M StreetView images.
SF-XL provides two query sets named \textit{test v1} and \textit{test v2}: (1) \textit{test v1} are 1000 images from Flickr, uniformly distributed across the whole city, providing some domain shifts (mostly viewpoint and a few night images); (2) \textit{test v2} is a set of 598 images taken with a smartphone in the city center, with images providing mild to moderate viewpoint shift \wrt to the database.
Although SF-XL large scale makes it a relevant option for research in VPR, its lack of well defined query splits from multiple domains makes it difficult to understand which methods can perform well in certain domain shifts.

To overcome the aforementioned limitations of previous datasets, we built two new sets of night and occluded images respectively, to be used against the database of SF-XL.

\myparagraph{Our new datasets}
\input{figures/queries_fig}
To obtain realistic and diverse queries, we downloaded hundreds of thousands images from Flickr for the area of San Francisco, similarly to \cite{Berton_2022_cosPlace, Radenovic_2019_gem, Philbin_2008_paris6k, Philbin_2007_oxford5k}.

With the help of trained classifiers, we then removed indoor images, and proceeded with the creation of two challenging sets of queries:
\begin{itemize}[noitemsep,topsep=1pt]
    \item \textit{SF-XL test night} is a set of night images, which we automatically selected with a trained classifier.
    \item \textit{SF-XL test occlusion} is a set of images that present heavy occlusions: the images were automatically selected using an object detection model, keeping those with a dynamic object (\eg car, truck, person) with width $>$ 50\% and height $>$ 30\% the size of the image.
\end{itemize}
Due to the inaccuracy of Flickr geotag information we manually verified the positions of each image, which resulted in 466 and 76 images for the \textit{SF test night} and \textit{SF test Occlusion} respectively.
A sample of queries is shown in \cref{fig:queries}.

Given the availability in literature of an open-source large-scale dataset that covers the city of San Francisco, namely SF-XL \cite{Berton_2022_cosPlace}, we match our proposed sets of queries against the SF-XL dataset, in practice using it as a database.


A summary of the datasets that we use in our benchmarks is shown in \cref{tab:dataset}.



%% file: figures/queries_fig.tex
\begin{figure}
    \centering
    \textbf{a) \ \ }
    \begin{minipage}{.17\textwidth}
        \includegraphics[width=\textwidth]{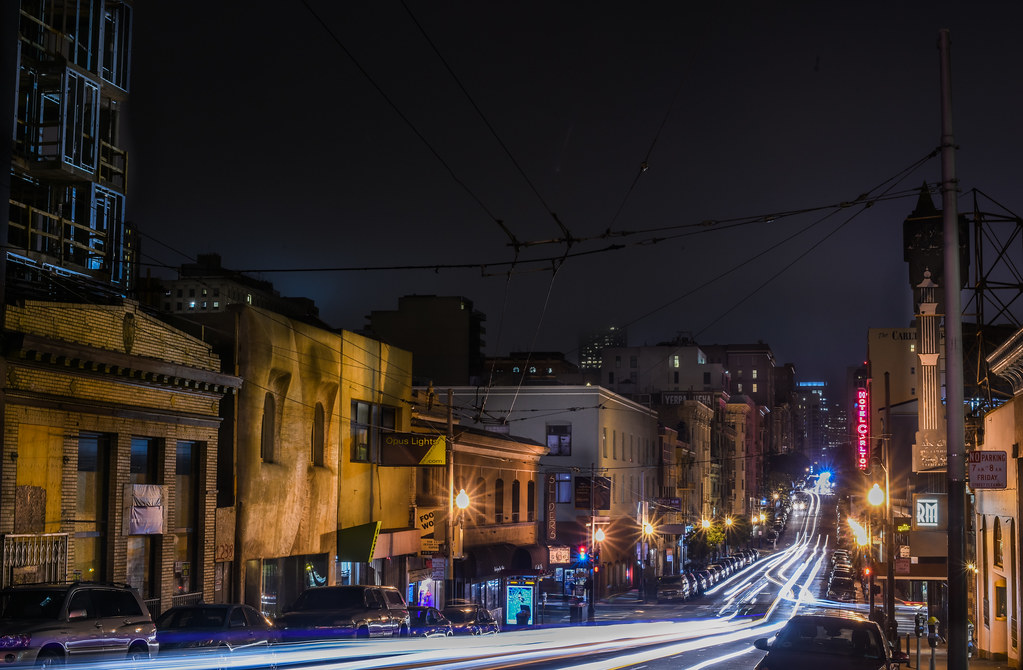}
    \end{minipage}
    \begin{minipage}{.12\textwidth}
        \includegraphics[width=\textwidth]{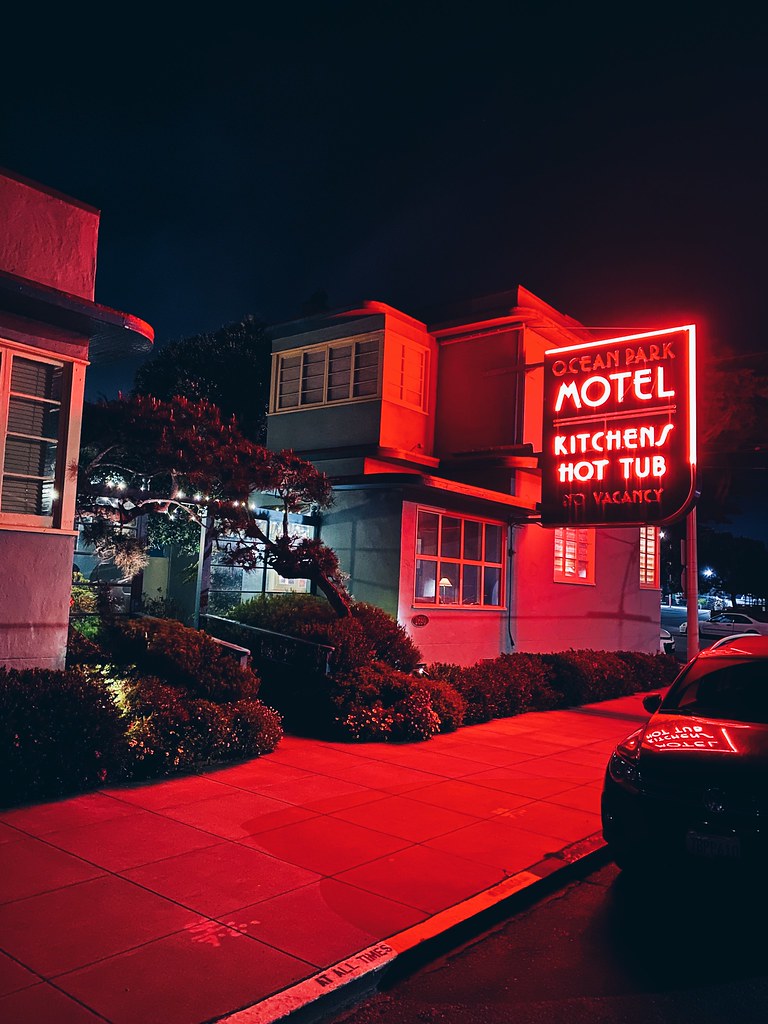}
    \end{minipage}
    \begin{minipage}{.12\textwidth}
        \includegraphics[width=\textwidth]{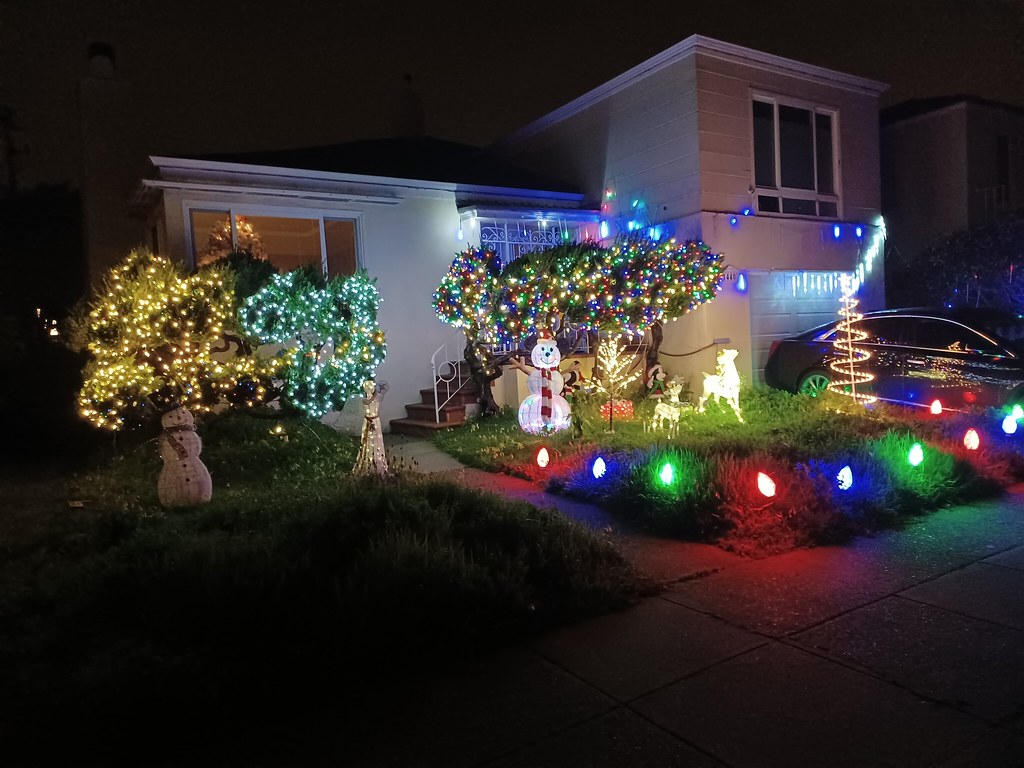}
    \end{minipage}
    \textbf{\\ b) }
    \begin{minipage}{.15\textwidth}
        \includegraphics[width=\textwidth]{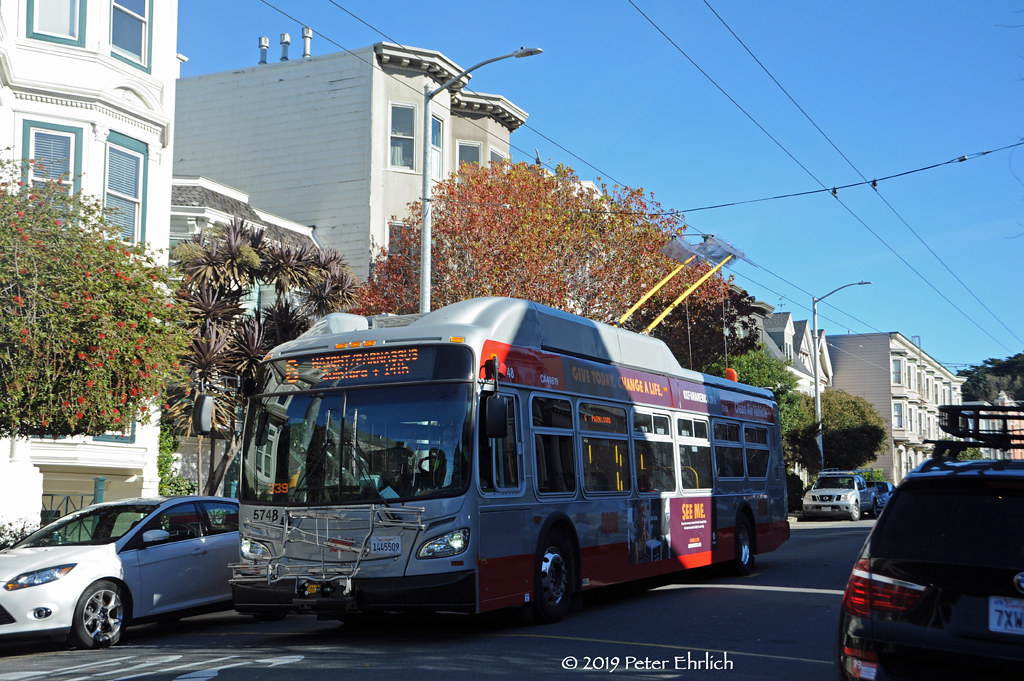}
    \end{minipage}
    \begin{minipage}{.12\textwidth}
        \includegraphics[width=\textwidth]{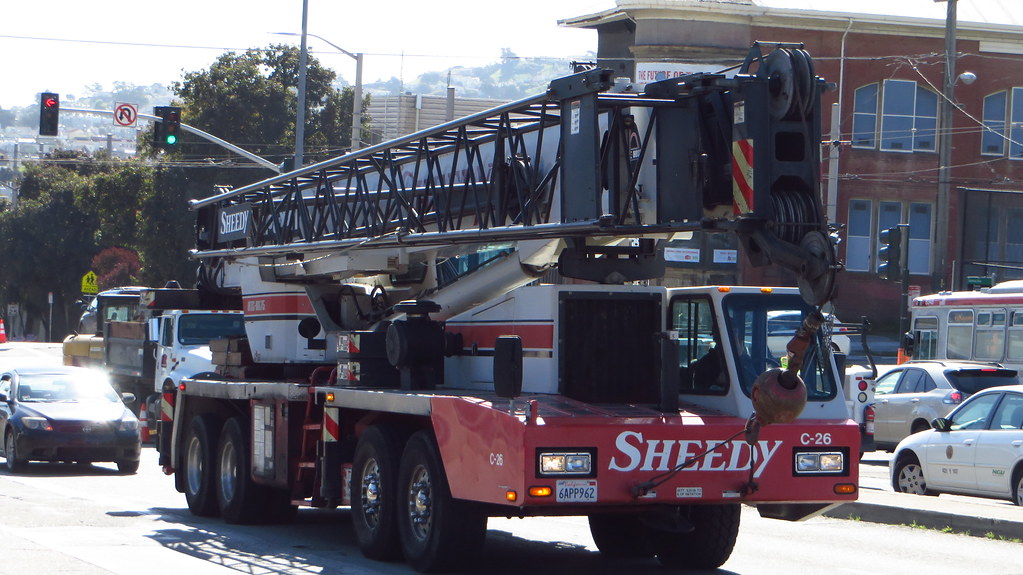}
    \end{minipage}
    \begin{minipage}{.15\textwidth}
        \includegraphics[width=\textwidth]{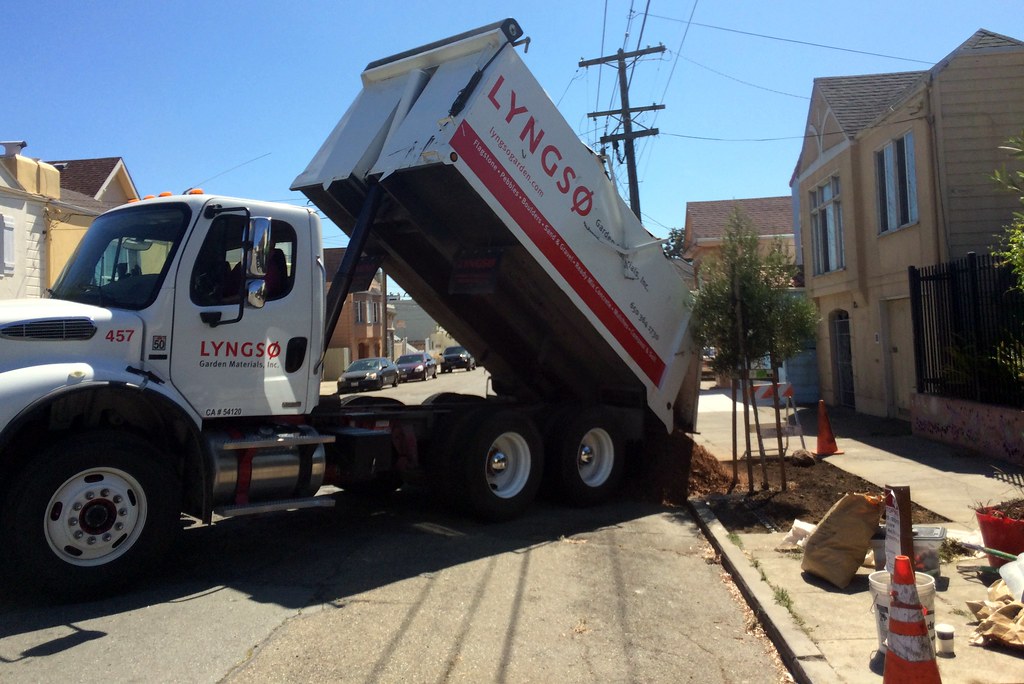}
    \end{minipage}
    \caption{\textbf{Examples of queries from a) SF-Night and b) Sf-occlusion.}}
    \label{fig:queries}
\vspace{-3mm}
\end{figure}

%% file: chapters/ch4_experiments.tex
\input{tables/global_descriptors_methods}

\section{Experiments}
\label{sec:experiments}
\input{tables/reranking_results}
\input{tables/characteristics_of_local_extractors}

\subsection{Benchmark Methodology}

To further motivate our benchmark, we point out that the application of spatial verification methods to the VPR task is not straightforward in light of the fact that many of them \cite{Sarlin_2020_superglue, Sun_2021_loftr} are trained on 3D models from SfM \cite{Li_2018_megadepth}, thus having access to accurate matching labels. On the other hand, in the place recognition settings,
matches are more loosely defined (within 25 m \cite{Arandjelovic_2018_netvlad, Berton_2022_benchmark, Pion_2020_benchmark}) and thus positive matches may share only a small portion of a scene. These differences raise doubts on the performances of these methods against complex perspective shifts and transient objects. In our proposed benchmark we shed a light on these previously unexplored research questions.

In our set of experiments, our aim is to maximize the results given the following two-step pipeline (see \cref{fig:reranking_pipeline}):
\begin{enumerate}[noitemsep,topsep=1pt]
    \item first we obtain a shortlist of K candidates using global descriptors methods (\ie the K nearest neighbor to the query in features space);
    \item sort the K candidates with a re-ranking algorithm.
\end{enumerate}
Given that a large body of literature on image retrieval through global descriptors already exists in the specific task of VPR \cite{Arandjelovic_2018_netvlad, Zaffar_2021_vprbench, Berton_2022_benchmark, Berton_2022_cosPlace, Alibey_2022_gsvcities, Alibey_2023_mixvpr}, our benchmark focuses on the second step, \ie the re-ranking algorithms.
To this end, we obtain a shortlist of candidates using CosPlace \cite{Berton_2022_cosPlace} (using a ResNet-50 backbone), which outperforms all other methods on SF-XL test v1 (see \cref{tab:global_descriptors_methods}).
Then, we perform the re-ranking step with a number of methods from the literature, namely SuperGlue, D2-Net, R2D2, DELG, Patch-NetVLAD, TransVPR, LoFTR and CVNet.
By providing these algorithms with the same set of candidates to re-rank, it is possible to disentangle the effect of the local features from the global extractor performance. In this way we obtain an indication of the benefit one can expect from the spatial verification step on this challenging task, quantifying the expected gains when modularly integrating these models a pre-existing VPR pipeline.

Given that re-ranking is performed on K candidates, the value of K is of great importance: higher values of K reduce the speed of the search, but (might) also lead to higher results (we investigate this effect in \cref{sec:ablation_K}).
For our main experiments we set $K=100$, following \cite{Hausler_2021_patch_netvlad, Wang_2022_TransVPR, Cao_2020_delg, Fuwen_2021_reranking_transformers}; in \cref{sec:ablation_K} we investigate how different values of K affect speed and results.

Following the VPR literature \cite{Arandjelovic_2018_netvlad, Ge_2020_sfrs, Liu_2019_sare, Zhu_2023_r2former, Berton_2022_cosPlace, Berton_2022_benchmark}, we use the Recall@N (R@N) as metric, which indicates the proportion of queries for which at least one of the first N predictions is correct, \ie within a given threshold distance from the query.
The threshold distance is set to 25 meters, although in \cref{fig:plots_different_K} we investigate how results change with a distance of 50 and 100 meters.
Note that a positive image might not share any visual content with the query (e.g. distance < 25 meters but opposite viewpoint direction with the query), although the chance that one of these positives is matched to a query by pure chance decreases as the database increases, and that it would otherwise be unfeasible to obtain unbiased ground-truth covisibility (as it would rely on one of the methods used for testing).


\subsection{Implementation details}

To provide a relevant benchmark, we use a large number of methods, some of which were specifically designed for re-ranking and some for tasks like spatial verification and image matching.
Specifically, we use SuperGlue \cite{Sarlin_2020_superglue} (which uses SuperPoint \cite{DeTone_2018_superpoint} local features), D2-Net \cite{Dusmanu_2019_D2Net}, R2D2 \cite{Revaud_2019_r2d2}, DELG \cite{Cao_2020_delg}, Reranking Transformers (RRT) \cite{Fuwen_2021_reranking_transformers} (which uses DELG local features), Patch-NetVLAD \cite{Hausler_2021_patch_netvlad} with both its RANSAC and Rapid Scoring implementation, TransVPR \cite{Wang_2022_TransVPR}, LoFTR \cite{Sun_2021_loftr} and CVNet \cite{Lee_2022_cvnet}.

For all methods, we use the official implementations and weights released by the authors, without fine-tuning. When more options were available we chose the configuration with best performance on \textit{SF-XL test v2}, with the only exception that we preferred models with ResNet50 backbone to ResNet100 counterpart. 

A number of methods use some kind of multi-scale approach, namely DELG, R2D2, D2-Net, Patch-NetVLAD and CVNet.
Most spatial verification methods rely on the standard RANSAC with 8 parameters to describe a homography, with the exceptions of DELG, which uses an affine version of RANSAC (\ie with 6 parameters).
Patch-NetVLAD applies RANSAC multiple times to match multiple scales (for the multi-scale approach).
Patch-NetVLAD also provides a fast version, which uses a novel alternative to RANSAC called Rapid Scoring, which is a non-iterative heuristics that has been proposed as a faster option to RANSAC as it does not require an iterative algorithm to be computed.
Preliminary experiments with Rapid Scoring on methods other than Patch-NetVLAD gave poor results, and we found it to be sensible to outliers and reliable only for moderate changes in viewpoint.
For Patch-NetVLAD and TransVPR we resized the images to have a resolution of 480x640, following the original implementations.


\subsection{Quantitative evaluations of results.}

Results from our experiments are shown in \cref{tab:reranking_results}.
We point out how the majority of the methods effectively provide a boost with respect to the baseline performance. For example, LoFTR and SuperGlue grant on average a 21\% and 13\% boost, respectively on night and occlusion benchmarks. This supports one of the motivation of our paper, that is showing the potential of methods based on local features to overcome the limitations of global descriptors in these challenging scenarios. We summarize other detailed findings from the experiment table in the following points:

\begin{itemize}[noitemsep,topsep=1pt]
    \item Interestingly, methods designed for Image Matching turn out to be highly competitive even when extended to VPR systems. In particular, SuperGlue achieves the best recalls for \textit{Tokyo night}, \textit{SF-XL test v1} and \textit{SF-XL test night}. This disproves the intuitive hypothesis that these models would suffer the absence of a training protocol that explicitly encodes a prior on ignoring transient objects. The same considerations hold as well for D2-Net and LoFTR, which perform very closely to SuperGlue across the board.
    
    \item Among the native re-ranking methods, DELG paired with RANSAC is the more versatile option. It reaches the best performance on \textit{SVOX night}, and on every other dataset its R@1 is comparable with the highest score. Regarding CVNet, it grants the highest R@5 and R@10 in almost every benchmark, despite some drops in R@1 for \textit{SVOX night} and \textit{SF-XL test v1}. TransVPR and Patch-NetVLAD end up being by far the less robust to the night domain.
    
    \item CVNet is the single best model on \textit{SF-XL test occlusion}. This confirm the effectiveness of its training procedure that involves Hide-and-Seek\cite{Singh_2017_hide_and_seek} augmentations for robustness against occlusions.

    \item Rapid scoring reaches very modest results, failing to provide a faster alternative for matching. Likewise, scoring with RRT improves the baseline but it is significantly worse than RANSAC variant on night datasets.
    
    \item Finally, it emerges how our newly proposed datasets are far from solved; we believe that this new and challenging benchmarks will inspire the community, paving the way for future research.
\end{itemize}


\subsection{Ablation on K and different positive threshold distances.}
\label{sec:ablation_K}
\input{figures/plots_different_K}

Given that re-ranking can be orders of magnitude more expensive than standard retrieval (through a nearest neighbor search), it is important to understand the ideal number of candidates to be re-ranked for an efficient VPR system.
To this end, we use the best-performing methods from \cref{tab:reranking_results} and run a new set of experiments by using different values of K, precisely any $K \in \{1, 2, 3, ... 100\}$.
Furthermore, we investigate how this affects the Recall@1 not only when using a threshold for positives of 25 meters, but also when increasing it to 50 and 100 meters.
Given the large number of combinations, we report the results on the two datasets that we believe would best represent a real-world scenario: \textit{SF-XL test v1} and \textit{SF-XL test night}.

Interestingly, we find that no single method achieves best results across the board:
\begin{itemize}[noitemsep,topsep=1pt]
  \item Firstly, it can be noted that simply increasing the threshold up to 100m allows to count more matches as correct. This is relevant as many applications that do not require high localization precision can exploit this effect. In particular, increasing the threshold grants higher gains on the more challenging datasets.
  \item DELG and CVNet exhibit superior performance when the threshold is increased to 100 meters: this is probably due to them being trained on the Google Landmark Dataset, which provides photos of buildings from far apart. In particular DELG scores higher on \textit{SF-XL test v1} whereas CVNet is superior in handling night images and transient occlusions.
  \item D2-Net, SuperGlue and DELG provide more precise matches under domain shift. They achieve the top scores with a threshold of 25 meters on \textit{SF-XL test night}.
  \item As a rule of thumb, the higher the K the better the results. However, it should be noted that the more challenging the dataset, the earlier this curve plateaus. This effect is especially visible with lower thresholds; in particular on \textit{SF-XL test occlusion} in many cases increasing K leads to higher false positives ratio. Considering that the cost of re-ranking scales linearly with K, the choice of this parameter must be devoted the utmost attention.

  \item Lastly, we can see that the upper bound (\ie the Recall@K with CosPlace) is still much higher than any of the re-ranking methods, proving that there is still a large margin for improvements.
\end{itemize}


\subsection{Qualitative evaluations of results.}
\input{figures/qualitatives}

To give an intuition to the reader over the strengths and weaknesses of three re-ranking methods (\ie SuperGlue, DELG and CVNet), we report in \cref{fig:qualitatives} a number of queries and the first prediction with each method.


\subsection{Is the night domain a real challenge?}

In this section we disentangle if the errors in night time datasets are due to the difficult illumination or other factors.
We considered the 101 queries of \textit{Tokyo night} for which CosPlace provides at least one positive within the first 100 candidates.
Of these 101 queries, we found that CVNet is able to solve every single query, while DELG and SuperGlue fail in one case (which is shown in \cref{fig:qualitatives} (d)).
Given these results, we argue that the difficulty of the \textit{SF-XL test night} dataset is not solely due to the night domain: for example, a factor could be that the night photos from Flickr  often contain several other challenges, representative of realistic use-cases. For instance heavy viewpoint shifts, and artificial lights such as signboards, decorative lights can highly affect the visual appearance of a place (\eg see queries in \cref{fig:queries}).
These results prove that state-of-the-art local features are indeed very robust to illumination changes, and that our newly proposed \textit{SF-XL test night} highlights the real challenges that photos in the wild can present.

\subsection{Computational cost}
In \cref{fig:plot_r1_time} we study the computational requirements for the considered re-ranking methods in relationship to their performances on \textit{SF-XL test night} and \textit{SF-XL test occlusion}. 
We plot the time required to re-rerank the top-100 candidates for a single query, considering online extraction of local features for the candidates from the database. Although many works consider this step to be computed offline, without any quantization techniques the storage cost would quickly explode on realistic large scale databases for VPR. For instance, storing SuperPoint local descriptors for the SF-XL database would require roughly 1 Tb. Since quantization techniques must be studied accurately for each case \cite{Noh_2017_delf, Berton_2022_benchmark}, we kept the most generally applicable implementation and considered online local features extraction. 
In general, methods with lighter backbones for feature extraction are the fastest, namely R2D2 and TransVPR. Whereas DELG, either with RANSAC or RRT, is the costlier approach. Including performances into the equation, SuperGlue, LoFTR and CVNet attain the best trade-off overall.
Nevertheless, these delays of the order of hundreds of seconds may not be acceptable in many practical applications, and this trade-off should be carefully evaluated together with the ablation on the number of candidates to re-rank presented in \cref{fig:plots_different_K}.
It shows that, despite it is common practice in the re-ranking literature to adopt $K=100$ or more \cite{Noh_2017_delf, Hausler_2021_patch_netvlad, Wang_2022_TransVPR, Cao_2020_delg, Fuwen_2021_reranking_transformers}, in many cases it is possible to cut down inference time substantially reducing K without suffering big performance hits.



%% file: tables/global_descriptors_methods.tex
\begin{table}
\centering
\begin{adjustbox}{width=0.7\columnwidth}
\begin{tabular}{lcccc}
\toprule
\multirow{2}{*}{\begin{tabular}[c]{@{}c@{}}Retrieval\\Method\end{tabular}} &
\multirow{2}{*}{\begin{tabular}[c]{@{}c@{}}Descriptors\\Dimension\end{tabular}} &
\multicolumn{3}{c}{SF-XL test v1} \\
\cline{3-5}
& & R@1 & R@5 & R@10 \\
\hline
NetVLAD  & 4096 & 33.1 & 45.0 & 50.4 \\
TransVPR & 256 &  9.7 & 16.6 & 20.3 \\
CVNet    & 2048 & 70.1 & 81.2 & 84.6 \\
DELG     & 2048 & 64.3 & 73.0 & 76.1 \\
CosPlace &  512 & \textbf{76.7} & \textbf{82.5} & \textbf{85.6} \\
Conv-AP  & 4096 & 49.1 & 60.6 & 65.6 \\
MixVPR   & 4096 & 72.3 & 79.5 & 81.4 \\
\bottomrule
\end{tabular}
\end{adjustbox}
\caption{\textbf{Recalls with different retrieval methods.} We used only global descriptors for this table (\ie no re-ranking is applied to DELG and CVNet). 
NetVLAD uses a VGG-16 \cite{Simonyan_2015_vgg} (and PCA), TransVPR a custom transformer model, while for all other methods we used the author's ResNet-50 \cite{He_2016_resnet} implementation.}
\label{tab:global_descriptors_methods}
\vspace{-3mm}
\end{table}

%% file: tables/reranking_results.tex
\begin{table*}
\begin{adjustbox}{width=\linewidth}
\centering
\begin{tabular}{llcccccccccccccccccccccccc}
\toprule
\multirow{3}{*}{{\begin{tabular}[c]{@{}c@{}}Features\\Extractor\end{tabular}}} & \multirow{3}{*}{{\begin{tabular}[c]{@{}c@{}}Features\\Matching\end{tabular}}} &
\multicolumn{3}{c}{Tokyo night} & & 
\multicolumn{3}{c}{SVOX night} & &
\multicolumn{3}{c}{SF-XL test v1} & &
\multicolumn{3}{c}{SF-XL test v2} & &
\multicolumn{3}{c}{SF-XL test night} & & 
\multicolumn{3}{c}{SF-XL test occlusion}  \\
& & 
\multicolumn{3}{c}{R@100 = 96.2} & & 
\multicolumn{3}{c}{R@100 = 90.3} & &
\multicolumn{3}{c}{R@100 = 92.5} & &
\multicolumn{3}{c}{R@100 = 97.7} & &
\multicolumn{3}{c}{R@100 = 41.6} & & 
\multicolumn{3}{c}{R@100 = 60.5}  \\
\cline{3-5} \cline{7-9} \cline{11-13} \cline{15-17} \cline{19-21} \cline{23-25}
 &  & R@1  & R@5  & R@10  & & R@1  & R@5  & R@10 & & R@1  & R@5  & R@10 & & R@1  & R@5  & R@10 & & R@1  & R@5  & R@10 & & R@1  & R@5  & R@10 \\
\hline

-                 & -          & 80.0 & 88.6 & 91.4 && 51.6 & 68.8 & 76.1 && 76.7 & 82.5 & 85.6 && 89.0 & 95.3 & 96.3 && 23.8 & 29.0 & 31.5 && 26.3 & 38.2 & 46.1  \\

SuperPoint    & SuperGlue      & \textbf{95.2} & \underline{95.2} & \underline{95.2} && 77.9 & \textbf{85.2} & \textbf{86.5} && \textbf{88.6} & \textbf{91.6} & \textbf{91.9} && 92.8 & \underline{96.7} & \textbf{97.7} && \textbf{33.0} & 38.0 & 39.1 && 38.2 & 44.7 & 50.0  \\
 
D2-net         & RANSAC        & 92.4 & \textbf{96.2} & \textbf{96.2} && 78.9 & \underline{85.1} & \underline{86.4} && 87.5 & 90.3 & 90.8 && \underline{94.0} & 96.3 & 97.0 && \underline{32.6} & \underline{38.2} & \underline{39.5} && \underline{40.8} & 48.7 & 51.3  \\

R2D2          & RANSAC         & 86.7 & 90.5 & 92.4 && 72.5 & 80.7 & 82.9 && 85.1 & 88.2 & 89.6 && \textbf{94.1} & \textbf{96.8} & 96.8 && 26.2 & 32.2 & 33.9 && 38.2 & 47.4 & 50.0  \\

DELG           & RANSAC        & \underline{94.3} & \underline{95.2} & \textbf{96.2} && \textbf{80.1} & 84.1 & 86.0 && \underline{88.5} & \underline{91.2} & 91.5 && 93.8 & 96.2 & 97.0 && 32.2 & 37.6 & 39.2 && 38.2 & \underline{50.0} & \underline{53.9} \\

DELG           & RRT           & 84.8 & 94.3 & \underline{95.2} && 66.3 & 81.7 & 85.7 && 85.3 & 89.6 & 90.4 && 88.6 & 96.0 & \underline{97.2} && 27.3 & 35.6 & 38.6 && 35.5 & 48.7 & 52.6 \\

Patch-NetVLAD  & RANSAC        & 90.5 & 94.3 & 94.3 && 67.2 & 80.6 & 83.6 && 77.0 & 84.7 & 87.0 && 91.0 & 95.2 & 96.2 && 31.8 & 37.3 & 38.4 && 34.2 & 47.4 & 52.6  \\

Patch-NetVLAD & Rapid Scoring  & 73.3 & 87.6 & 92.4 && 42.2 & 66.3 & 73.1 && 69.3 & 80.3 & 84.1 && 90.0 & 94.6 & 95.8 && 21.7 & 31.3 & 35.4 && 25.0 & 38.2 & 42.1 \\

TransVPR      & RANSAC         & 88.6 & \underline{95.2} & \underline{95.2} && 63.8 & 79.2 & 83.2 && 84.0 & 87.6 & 89.1 && 92.5 & 96.2 & 96.7 && 27.3 & 34.3 & 36.7 && 38.2 & 46.1 & 52.6 \\

\multicolumn{2}{c}{LoFTR}      & 93.3 & \underline{95.2} & \underline{95.2} && \underline{80.0} & 84.0 & 85.3 && 87.9 & 89.8 & 90.7 && 93.3 & 96.3 & \underline{97.2} && \underline{32.6} & 37.6 & 38.2 && \underline{40.8} & 48.7 & 51.3 \\

\multicolumn{2}{c}{CVNet}      & \underline{94.3} & \textbf{96.2} & \textbf{96.2} && 74.6 & \textbf{85.2} & \textbf{86.5} && 84.8 & 91.0 & \underline{91.6} && 88.0 & 95.8 & 97.0 && 31.5 & \textbf{39.3} & \textbf{39.9} && \textbf{42.1} & \textbf{52.6} & \textbf{56.6}  \\

\bottomrule
\end{tabular}
\end{adjustbox}
\caption{\textbf{Recalls before and after applying re-ranking.} 
The shortlist of candidates to be re-ranked is obtained with CosPlace, and the results with such shortlist are shown in the first row.
Re-ranking has been applied to the first 100 candidates (\ie $K=100$). Next to each dataset's name, we show the R@100, which in practice sets the upper bound of the maximum recalls achievable after re-ranking. Best results are in \textbf{bold}, second best are \underline{underlined}.
}
\label{tab:reranking_results}
\end{table*}

%% file: tables/characteristics_of_local_extractors.tex
\begin{table}[!t]
\centering
\begin{adjustbox}{width=\linewidth}
\begin{tabular}{lccccc}
\toprule
\multirow{2}{*}{}{Model} & 

\multirow{2}{*}{}{\begin{tabular}[c]{@{}c@{}}Descriptors size\\(num. $\times$ dim.)\end{tabular}} & 
\multirow{2}{*}{}{\begin{tabular}[c]{@{}c@{}}Backbone\end{tabular}} &
\multirow{2}{*}{}{\begin{tabular}[c]{@{}c@{}}Designed for\\re-ranking\end{tabular}} &
\multirow{2}{*}{}{\begin{tabular}[c]{@{}c@{}}Sparse\\Keypoints\end{tabular}} \\  
\midrule
DELG          & 1000 x 128  & ResNet-50 &\GT&\GT\\
Patch-NetVLAD & 2826 x 4096 & VGG-16 &\GT&\RX\\
TransVPR      & 522 x 256   & Custom CNN+transformer &\GT&\GT\\
R2D2          & 4126 x 128  & custom L2-Net\cite{Tian_2017_l2net} &\RX&\GT\\
D2Net         & 2775 x 512  & VGG-16 &\RX&\GT\\
SuperPoint    & 1034 x 256  & custom VGG &\RX&\GT\\
\bottomrule
\end{tabular}
\end{adjustbox}
\caption{\textbf{Characteristics of local features extractors.}
The descriptors size was computed for all methods on the same image of resolution 480x640. For Patch-NetVLAD descriptors size depends only on the resolution, because it uses dense keypoints/features, whereas 
for all other methods the number of descriptors depends on the visual content of the image.
}
\label{tab:characteristics_of_local_extractors}
\vspace{-3mm}
\end{table}

%% file: figures/plots_different_K.tex
 
\begin{figure*}
    \centering
    \includegraphics[width=\textwidth]{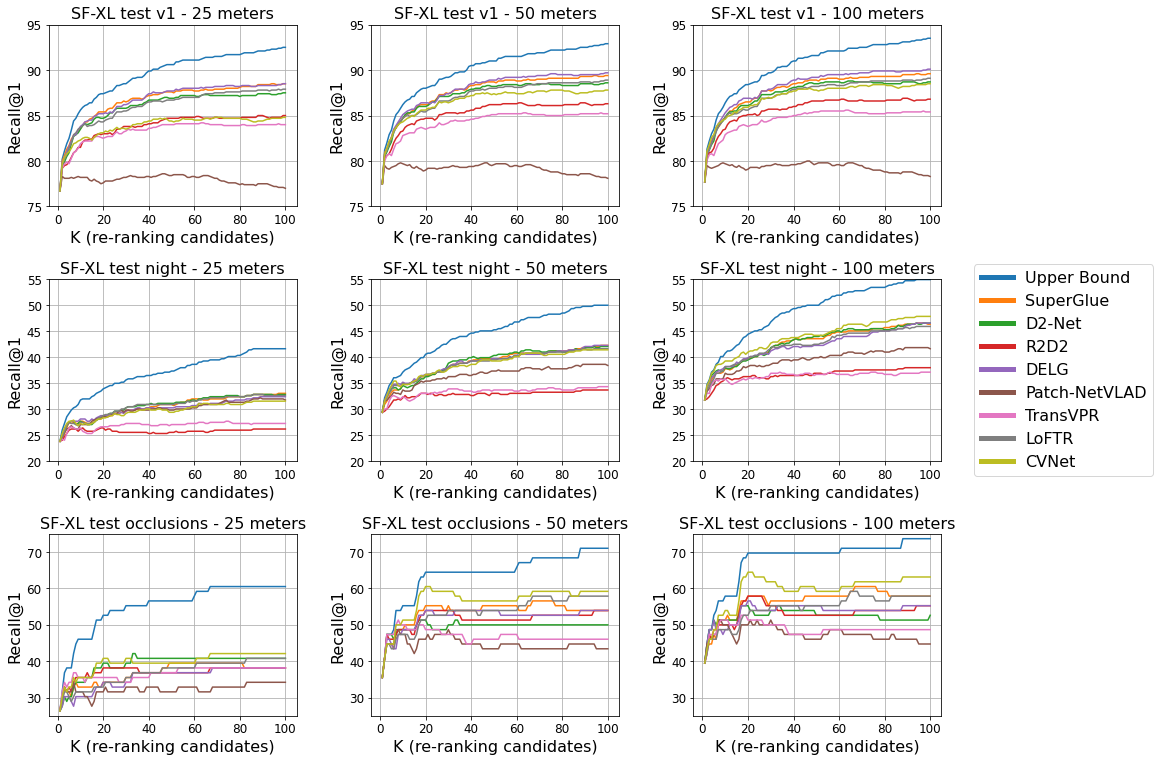}
    \caption{\textbf{Re-ranking with different values of K, from 1 to 100.} The "Upper Bound" is the Recall@K without applying re-ranking (\ie with CosPlace). For DELG and Patch-NetVLAD we used the version with RANSAC.}
    \label{fig:plots_different_K}
\end{figure*}

%% file: figures/qualitatives.tex
\begin{figure*}
    \centering
    \begin{minipage}{0.99\linewidth}
        \includegraphics[width=\linewidth]{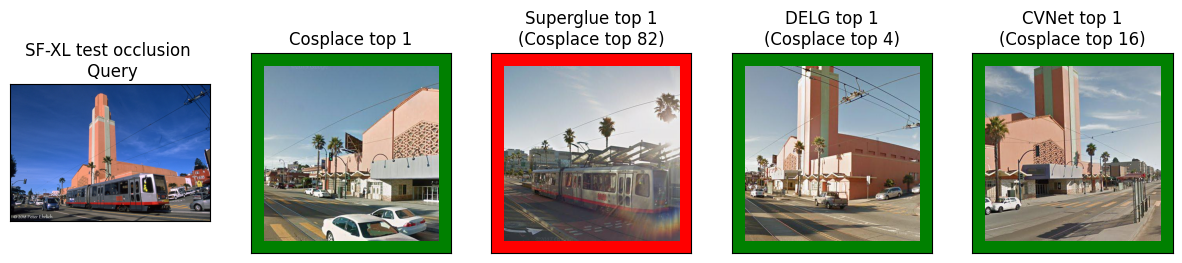}
        \subcaption[]{A failure of SuperGlue due to a dynamic object (a tram), which SuperGlue (unlike DELG and CVNet) has not been trained to ignore. We can also see that CVNet finds a positive with very different viewpoint than the query, even though candidates closer to the query are available.}
    \end{minipage}
    \begin{minipage}{0.99\linewidth}
        \includegraphics[width=\linewidth]{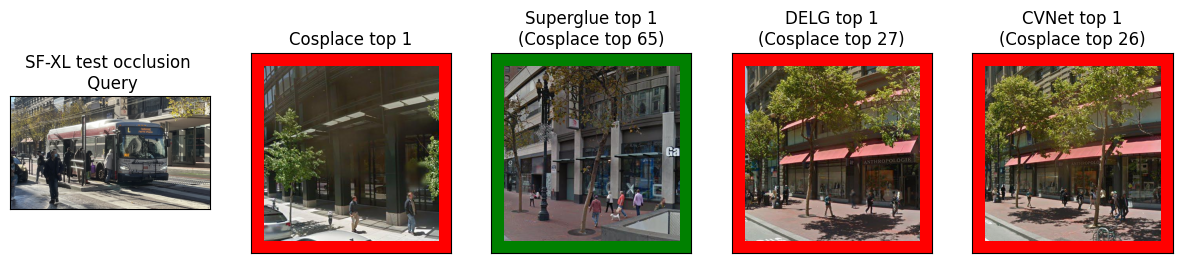}
    \subcaption[]{DELG and CVNet failures for this case are most likely due to those methods using a combination of local and global scoring system. The global features see trees and a red line (which for the query is on the bus, and for the predictions is an awning).}
    \end{minipage}
    \begin{minipage}{0.99\linewidth}
        \includegraphics[width=\linewidth]{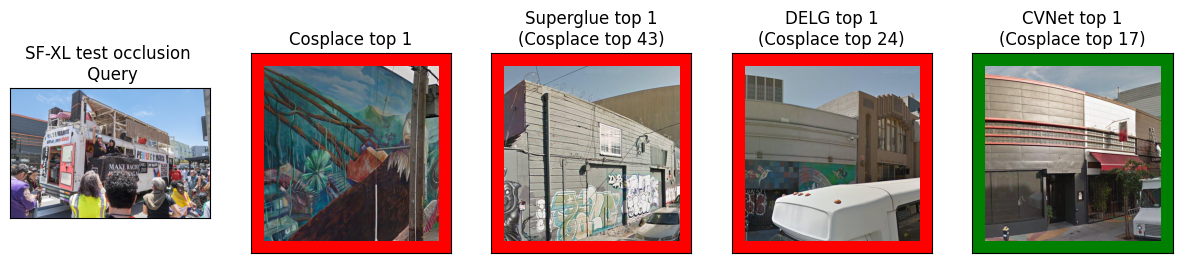}
        \subcaption[]{This example shows the robustness of CVNet to strong occlusions, which is learned thanks to its use of Hide-and-Seek data augmentation \cite{Singh_2017_hide_and_seek}.}
    \end{minipage}
    \begin{minipage}{0.99\linewidth}
        \includegraphics[width=\linewidth]{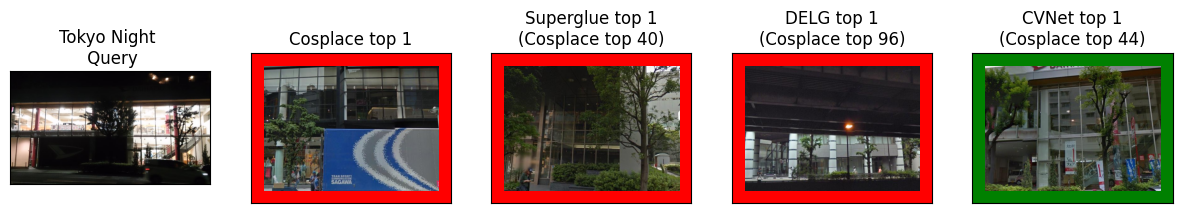}
        \subcaption[]{The only example from \textit{Tokyo night} where DELG and SuperGlue fail to find the correct prediction.}
    \end{minipage}
    \caption{\textbf{Qualitative examples of 3 queries and the first prediction with 4 relevant methods}, namely CosPlace (retrieval baseline) SuperGlue, DELG and CVNet. Predictions are in green if they are less than 100 meters away from the query's ground truth.}
    \label{fig:qualitatives}
\end{figure*}

%% file: chapters/ch5_conclusion.tex
\section{Conclusion}
\label{sec:conclusion}

In this paper we investigate how re-ranking techniques can be used to improved results in visual place recognition.
Specifically, we experiment on the relevant setting when the queries come from a different domain than the database, with a focus on night and occluded queries.
We propose two challenging query sets, on which even the best combination of methods achieve a Recall@1 $< 50\%$.

We provide a large set of experiments to show which methods perform best for the task of cross-domain re-ranking for VPR, finding that many methods achieve pareto-optimal solutions when time and recalls are considered.
We also find that different domain shifts require different approaches, and that there is no clear winner across all datasets, even when latency is not an issue.

We believe that our work can shed light on how to design a highly performant VPR system on multiple conditions, and that our proposed datasets will foster further research to continue to improve the state of the art.

\noindent\textbf{Acknowledgements.}
\small{This work was supported by CINI.}